\documentclass[journal, onecolumn,11pt]{IEEEtran}
\IEEEoverridecommandlockouts
\usepackage{cite}
\usepackage{amsmath,amssymb,amsfonts}
\usepackage{algorithm}
\usepackage{algpseudocode}
\usepackage{graphicx}
\usepackage{textcomp}
\usepackage{xcolor}
\usepackage{diagbox}
\usepackage{booktabs}
\usepackage{multirow}
\usepackage{float}

\usepackage[table]{xcolor} % Add this line in the preamble
\usepackage{hyperref}
\definecolor{darkgreen}{rgb}{0.0,0.45,0.0}  % tweak the RGB if you’d like it 
\definecolor{darkyellow}{rgb}{0.7,0.55,0}  % tweak values to taste
\definecolor{lightblue}{rgb}{0.20,0.50,0.90}  % adjust the RGB values if you want it even lighter/darker
\definecolor{darkred}{rgb}{0.55,0.0,0.0}  % deepen or lighten by tweaking the first number

\definecolor{lightred}{rgb}{1,0.4,0.4}    % soften/brighten by nudging the 0.4 values

\definecolor{vinotinto}{RGB}{109, 10, 31}    % deep maroon (≈ #6D0A1F); tweak as you like
\definecolor{midnightblue}{rgb}{0.05,0.20,0.55}
\definecolor{slategray}{rgb}{0.30,0.36,0.42}
\definecolor{charcoal}{rgb}{0.20,0.22,0.25}

\definecolor{burntsienna}{rgb}{0.74,0.33,0.16}
\definecolor{copper}{rgb}{0.72,0.45,0.20}
\definecolor{marigold}{rgb}{0.93,0.56,0.00}
\definecolor{darkmarigold}{rgb}{0.79, 0.48, 0.00}

\definecolor{turquoise}{rgb}{0.00,0.60,0.66}
\definecolor{indigo}{rgb}{0.29,0.00,0.51}
\definecolor{forestgreen}{rgb}{0.13,0.55,0.13}

\hypersetup{
    colorlinks = true,
    linkcolor  = darkgreen,
    filecolor  = darkgreen,
    urlcolor   = magenta,
    citecolor  = darkgreen,
}

\usepackage{balance}

\def\BibTeX{{\rm B\kern-.05em{\sc i\kern-.025em b}\kern-.08em
    T\kern-.1667em\lower.7ex\hbox{E}\kern-.125emX}}
\begin{document}

\title{DICE: Diffusion Consensus Equilibrium for Sparse-view CT Reconstruction \\

\thanks{This work was supported by the Agencia Nacional de Hidrocarburos (ANH) and the Ministerio de Ciencia, Tecnología e Innovación (MINCIENCIAS), under contract 045-2025. \\

© 2025 IEEE. Personal use of this material is permitted. Permission from IEEE must be obtained for all other uses, in any current or future media, including reprinting/republishing this material for advertising or promotional purposes, creating new collective works, for resale or redistribution to servers or lists, or reuse of any copyrighted component of this work in other works.

}
}
\author{
\IEEEauthorblockN{Leon Suarez-Rodriguez$^{\ddagger}$, Roman Jacome$^\dag$, Romario Gualdrón-Hurtado$^{\ddagger}$, Ana Mantilla-Dulcey$^{\S}$, Henry Arguello$^{\ddagger}$}
\IEEEauthorblockA{
$^{\ddagger}$Department of Systems and Informatics Engineering
\\
$^{\dagger}$Department of Electrical, Electronics, and Telecommunications Engineering
\\
$^{\S}$Department of Physics
\\
Universidad Industrial de Santander, Bucaramanga, Colombia, 680002}
}

\maketitle

\begin{abstract}

Sparse-view computed tomography (CT) reconstruction is fundamentally challenging due to undersampling, leading to an ill-posed inverse problem. Traditional iterative methods incorporate handcrafted or learned priors to regularize the solution but struggle to capture the complex structures present in medical images. In contrast, diffusion models (DMs) have recently emerged as powerful generative priors that can accurately model complex image distributions. In this work, we introduce Diffusion Consensus Equilibrium (DICE), a framework that integrates a two-agent consensus equilibrium into the sampling process of a DM. DICE alternates between: (i) a data-consistency agent, implemented through a proximal operator enforcing measurement consistency, and (ii) a prior agent, realized by a DM performing a clean image estimation at each sampling step. By balancing these two complementary agents iteratively, DICE effectively combines strong generative prior capabilities with measurement consistency. Experimental results show that DICE significantly outperforms state-of-the-art baselines in reconstructing high-quality CT images under uniform and non-uniform sparse-view settings of 15, 30, and 60 views (out of a total of 180), demonstrating both its effectiveness and robustness.

\end{abstract}


%\vspace{-0.1cm}
\section{Introduction}

Computed tomography (CT) is a widely used non-invasive medical imaging technique where acquiring high-quality reconstructions is essential for accurate clinical diagnosis. However, obtaining such images typically demands a large number of X-ray projections, increasing both radiation exposure and scan duration \cite{CT_IMPORTANCE}. To mitigate these limitations, sparse-view CT has been adopted, significantly reducing the number of required projections \cite{sparse_CT_paper_2002}. Formally, the sparse-view CT acquisition process can be represented as:
%\vspace{-0.2cm}
\begin{equation}
%\vspace{-0.2cm}
\boldsymbol{y}=\boldsymbol{A}\boldsymbol{x} + \boldsymbol{e},\label{eq:measurements}
\end{equation}

\noindent where $\boldsymbol{y} \in \mathbb{R}^m$ is the undersampled sinogram, $\boldsymbol{x} \in \mathbb{R}^n$ is the image to be reconstructed, $\boldsymbol{A} \in \mathbb{R}^{m\times n}$ is the forward operator corresponding to an undersampled Radon transform, and $\boldsymbol{e}$ represents acquisition noise. Sparse sampling inherently makes the inverse problem severely ill-posed, resulting in infinitely many solutions consistent with the observed data 
\cite{sparse_CT_intro_paper}.

To address this issue, iterative methods formulate CT recovery as a regularized optimization problem \cite{TV_CT, INVERSE_PROBLEM_OPT, PnP_paper}, explicitly balancing data fidelity with handcrafted or learned priors:
%\vspace{-0.1cm}
\begin{equation}
%\vspace{-0.1cm}
    \boldsymbol{x}^* = \underset{\boldsymbol{x}}{\mathrm{argmin}}\{f(\boldsymbol{x})+ \lambda h(\boldsymbol{x})\},
\end{equation}

\noindent where $f(\boldsymbol{x})$ enforces measurement consistency and $h(\boldsymbol{x})$ is a regularizer that incorporates prior information about the image manifold, and $\lambda$ is a regularization parameter. Although such approaches can significantly enhance image quality, they are limited by their ability to model complex medical image distributions accurately \cite{NEURAL_NETS_IP, NEURIPS2021_DM_MRI}.

Recent advancements in deep learning have further improved reconstruction quality by incorporating powerful data-driven priors. These methods typically fall into two categories: (i) plug-and-play (PnP) approaches \cite{PnP_paper, PnP_CHAN, PnP_Deep_Denoiser}, which integrate pretrained denoisers into iterative schemes, and (ii) fully end-to-end trained neural networks requiring paired datasets of undersampled measurements and corresponding clean images \cite{DCNN_CT, UNET_CT, DENSENET_CT}. While successful, these strategies rely on extensive training datasets or require significant retraining for variations in sampling scenarios or noise conditions \cite{DCNN_CT, LEARN_CT, DRONE_CT, HYBRID_CT}.

Diffusion models (DMs) \cite{Diffusion_dickstein15, DDPM, song2021denoising, song2021scorebased} have recently gained attention due to their exceptional capability in modeling complex image distributions via an iterative noising-denoising process. Conditioning DMs entails guiding their generative reverse diffusion process using measured data to ensure reconstructions align with the measurements, leading to substantial performance improvements across diverse inverse problems, including seismic imaging \cite{Paul_Diffusion, luismi_diffusion}, magnetic resonance imaging \cite{NEURIPS2021_DM_MRI, SCORE_MRI}, and classical image restoration tasks such as deblurring, inpainting, and super-resolution \cite{DPS, DiffPIR, DDRM}. However, conditioning these models introduces the challenge of properly balancing data consistency with perceptual realism \cite{RESAMPLE, DeppDataConsistency, Perception_distortion}.

To address this trade-off, we introduce Diffusion Consensus Equilibrium (DICE), a principled reconstruction framework that combines pretrained diffusion priors with measurement consistency using the consensus equilibrium (CE) framework. CE \cite{ConcensusEquilibriumPaper, ComputationalImagingBookBouman, Consensus_PAUL} is an optimization-free framework that generalizes traditional regularized inversion by defining reconstruction as an equilibrium between multiple independent agents, each promoting distinct reconstruction characteristics. In DICE, we formulate CT reconstruction as the CE problem of two agents during the sampling process of a DM. Specifically,  between:
(i) a data-consistency agent, realized by a proximal operator enforcing measurement consistency; and
(ii) a diffusion agent, implemented as a pretrained DM performing a clean image estimation at each sampling iteration. This integration effectively balances data consistency and the powerful generative capabilities of diffusion priors, iteratively driving the solution toward high-quality reconstructions. Experiments demonstrate that DICE outperforms state-of-the-art baselines in uniform and non-uniform sparse-view CT scenarios, achieving superior image quality under severe undersampling conditions.

%%%%%%%%%%%%%%%%%%%%%%%%%%%%%%%%

\section{DICE: Diffusion Consensus equilibrium}

The goal of DICE is to recover an image $\boldsymbol{x}^*$ that is simultaneously (i) consistent with the measured projections \eqref{eq:measurements} and (ii) realistic concerning the prior distribution learned by a well-trained DM $p(\boldsymbol{x}^*)$. For this, we adopt the CE formulation, which enables the integration of heterogeneous maps/agents, such as data-fidelity terms, proximal operators, or learned networks, and seeks an equilibrium point at which all these maps agree \cite{ConcensusEquilibriumPaper, ComputationalImagingBookBouman}.

In particular, we consider two agents: (i) a proximal operator $F_1:\mathbb{R}^n\rightarrow\mathbb{R}^n$ that promotes measurement consistency, and (ii) a generative prior $F_2:\mathbb{R}^n\rightarrow\mathbb{R}^n$ that produces a clean image estimation $\boldsymbol{x}_{0|t}$ at a timestep $t\in\{T,\dots,1\}$. These two agents are defined as follows:

%\vspace{-0.1cm}
\begin{equation}
    \begin{cases}
         F_1(\boldsymbol{v}_1)=\underset{\boldsymbol{s}}{\mathrm{argmin}} \left\{ \ \frac{1}{2} \| \boldsymbol{A}\boldsymbol{s}-\boldsymbol{y}\|_2^2 + \frac{\zeta_t}{2}\|\boldsymbol{s}-\boldsymbol{v}_1\|_2^2 \right\}, \\
         F_2(\boldsymbol{v}_2)=\frac{1}{\sqrt{\bar{\alpha}_t}}(\boldsymbol{v}_2 - \epsilon_{\boldsymbol{\theta}}(\boldsymbol{v}_2,t)\sqrt{1-\bar{\alpha}_t}),
    \end{cases}
    \label{eq:forces}
\end{equation}

\noindent where $\zeta_{t}=(1-\bar{\alpha}_{t})/\bar{\alpha}_{t}$ is a time-dependent penalty that links the CE updates to the DM noise schedule  $\{\beta_t\}_{t=1}^{T}\subset(0,1)$, with $\bar{\alpha}_t=\prod_{i=0}^t \alpha_i$, $\alpha_t = 1 - \beta_t$, and $\epsilon_{\boldsymbol{\theta}}$ is the DM's noise estimation network \cite{DDPM}.

Intuitively, agent $F_1$ tries to pull the solution to an image that is consistent with the measurements $\boldsymbol{y}$. In contrast, agent $F_2$ pulls the solution to a denoised or clean image $\boldsymbol{x}_{0|t}$ at a given timestep. Balancing these two agents at a timestep $t$ produces a clean image $\boldsymbol{x}_{0|t}^*$ that is consistent with the measurements. This image is obtained by finding a equilibrium point $(\boldsymbol{x}_{0|t}^*, \mathbf{u}^*)\in \mathbb{R}^n\times \mathbb{R}^{2n}$ such that: 
\begin{equation}
\begin{cases}
    F_1\!\left(\boldsymbol{x}_{0|t}^*+\boldsymbol{u}_1^*\right)=\boldsymbol{x}_{0|t}^*,\\
     F_2\!\left(\boldsymbol{x}_{0|t}^*+\boldsymbol{u}_2^*\right)=\boldsymbol{x}_{0|t}^* \\
     \bar{\boldsymbol{u}}_{\tau}^*=\tau_1 \boldsymbol{u}_1^* + \tau_2 \boldsymbol{u}_2=\boldsymbol{0},
\end{cases}
\label{eq:CE}
\end{equation}

\noindent where $\boldsymbol{u}_i \in \mathbb{R}^n$ is an offset variable of the agent $F_i$, $\mathbf{u}=\left(\boldsymbol{u}_1^\top, \boldsymbol{u}_2^\top\right)^\top \in \mathbb{R}^{2n}$ is the stacked offset variables, and  $\tau_i>0$ are positive weights which represent the relative strength of each agent $F_i$, with $\tau_1+\tau_2 =1$. In Eq. \eqref{eq:CE}, each agent $F_i$ can be interpreted as a force that moves the state $\boldsymbol{x}_{0|t}+\boldsymbol{u}_i$ toward the common
consensus point $\boldsymbol{x}_{0|t}^*$ where all agents agree \cite{ComputationalImagingBookBouman}. 

Introducing the change of variable for the state $\boldsymbol{v}_i=\boldsymbol{x}_{0|t}+\boldsymbol{u}_i$ with $\boldsymbol{v}_i\in \mathbb{R}^n$, we obtain:

\begin{equation}
    F_1(\boldsymbol{v}_1^*)=F_2(\boldsymbol{v}_2^*)=\boldsymbol{x}_{0|t}^*.
    \label{eq:change_variable}
\end{equation}

Let's define the stacked operator of agents $\mathbf{F}: \mathbb{R}^{2n}\rightarrow \mathbb{R}^{2n}$ and stacked weighted averaging-operator $\mathbf{G}_{\tau}: \mathbb{R}^{2n}\rightarrow \mathbb{R}^{2n}$:

\begin{equation}
    \mathbf{F}(\mathbf{v}) \;=\;
    \left(
    \begin{array}{c}
    F_{1}(\boldsymbol{v}_{1})\\
    F_{2}(\boldsymbol{v}_{2})
    \end{array}
    \right)
    \quad\text{and}\quad
    \mathbf{G}_{\tau}(\mathbf{v}) \;=\;
    \left(
    \begin{array}{c}
    \bar{\boldsymbol{v}}_{\tau}\\
    \bar{\boldsymbol{v}}_{\tau}
    \end{array}
    \right),
\end{equation}

\noindent where $\mathbf{v}=\left(\boldsymbol{v}_1^\top,  \boldsymbol{v}_2^\top\right)^\top \in \mathbb{R}^{2n}$ is the stacked vector of states, and ${\bar{\boldsymbol{v}}_{\tau}}$ is the weighted average vector of states, with  ${\bar{\boldsymbol{v}}_{\tau}=\sum_{i=1}^{N}\tau_i\boldsymbol{v}_i}$.

Let $\hat{\mathbf x}_{0|t}\in\mathbb{R}^{2n}$ contain two stacked copies of a vector
$\boldsymbol{x}_{0|t}\in\mathbb{R}^n$. With these definitions, Eq. \eqref{eq:CE} is equivalent to the system:

\begin{equation}
\mathbf{F}\!\left(\hat{\mathbf{x}}_{0|t}^*+\mathbf{u}^{*}\right)=\hat{\mathbf{x}}_{0|t}^*,
\qquad \bar{\mathbf u}_{\tau}^*=\boldsymbol{0},
\label{eq:CE_system}
\end{equation}

\noindent where $\bar{\mathbf u}_{\tau}^*=\left((\bar{\boldsymbol{u}}_{\tau}^*)^\top,  (\bar{\boldsymbol{u}}_{\tau}^*)^\top\right)^\top \in \mathbb{R}^{2n}$ is the vector of stacked copies of $\bar{\boldsymbol{u}}_{\tau}^*$. A point $(\boldsymbol{x}_{0|t}^*, \mathbf{u}^*)$ is the solution of Eq. \eqref{eq:CE_system} if only if $\mathbf{v}^*=\hat{\mathbf{x}}_{0|t}^*+\mathbf{u}^*$ satisfies:

\begin{equation}
    \begin{cases}
        \bar{\mathbf{v}}_\tau^*=\boldsymbol{x}_{0|t}^*, \\ \mathbf{F}(\mathbf{v}^*)=\mathbf{G}_\tau (\mathbf{v}^*) \\
        (2\mathbf{G}_\tau-\mathbf{I})(2\mathbf{F}-\mathbf{I})\mathbf{v}^*=\mathbf{v}^*
    \end{cases}
\end{equation}

\begin{algorithm}[t!]
    \caption{\textbf{DICE:} Diffusion Consensus Equilibrium}
    \label{alg:dice}
    \begin{algorithmic}[1]
        \Require DM noise estimator network $\epsilon_{\boldsymbol{\theta}}$, measurements $\boldsymbol{y}$, forward model $\boldsymbol{A}$, conjugate gradient descent steps $P$, CE agents $F_{1} (\boldsymbol{y}, \boldsymbol{A}, P)$, and $F_{2}(\epsilon_{\boldsymbol{\theta}})$, relaxation parameter $\rho\!\in\!(0,1)$, number of Mann iterations $K$, and noise schedule $\{\beta_{t}\}_{t=1}^{T}$
        \State $\boldsymbol{x}_{T}\sim\mathcal{N}(\mathbf 0,\boldsymbol{I})$ \Comment{Latent initialization}
        \For{$t=T,\dots,1$}    \Comment{Reverse‑diffusion steps}
            
            \State $\mathbf{v}\gets \begin{bmatrix}
                \boldsymbol{x}_{t} \\
                \boldsymbol{x}_{t}
            \end{bmatrix}$            \Comment{Initial CE state}
            \For{$k=1,\dots,K$}               \Comment{Mann iteration}
                \State $\mathbf{v}'\gets (2\mathbf{G}-\mathbf{I})(2\mathbf{F}-\mathbf{I})\,\mathbf{v}$
                \State $\mathbf{v}\gets(1-\rho)\,\mathbf{v}+\rho\,\mathbf{v}'$
            \EndFor

            \State $\boldsymbol{x}_{0|t}^*\gets \tau_1 \boldsymbol{v}_1^*+\tau_2\boldsymbol{v}_2^*$
             
            \State $\tilde{\boldsymbol{x}}_{t-1}=\sqrt{\bar{\alpha}_{t-1}} \boldsymbol{x}_{0|t}^* + \sqrt{1-\bar{\alpha}_{t-1}}\boldsymbol{z}$ \Comment{$\boldsymbol{z}\sim \mathcal{N}(0,\boldsymbol{I})$}
            
        \EndFor

        \State
        $\boldsymbol{x}^* \gets \tilde{\boldsymbol{x}}_0$
        \State \textbf{return} $\boldsymbol{x}^*$
    \end{algorithmic}
    
\end{algorithm}

This means that $\mathbf{v}^*$ is a fixed point of $\mathbf{\Omega}=(2\mathbf{G}_\tau-\mathbf{I})(2\mathbf{F}-\mathbf{I})$ \cite{ConcensusEquilibriumPaper, ComputationalImagingBookBouman}. By  using Mann iterations, we can find the fixed point of $\mathbf{\Omega}$ as follows:

\begin{equation}
    \mathbf{v}^{j+1}=(1-\rho)\mathbf{v}^j+\rho \mathbf{\Omega}(\mathbf{v}^j) \qquad
  0<\rho<1.
\end{equation}

Once $\mathbf{v}^*=\left((\boldsymbol{v}_1^*)^\top, (\boldsymbol{v}_2^*)^\top\right)^\top$ is reached, the consensus solution is  $\boldsymbol{x}_{0|t}^*= \tau_1 \boldsymbol{v}_1^* + \tau_2 \boldsymbol{v}_2^*$.

Then, to obtain the next sample step $\tilde{\boldsymbol{x}}_{t-1}$, we introduce $\boldsymbol{x}_{0|t}^*$ back to the diffusion process by performing the next forward diffusion step:

\begin{equation}
    \tilde{\boldsymbol{x}}_{t-1}=\sqrt{\bar{\alpha}_{t-1}} \boldsymbol{x}_{0|t}^* + \sqrt{1-\bar{\alpha}_{t-1}}\boldsymbol{z}, \quad \boldsymbol{z}\sim \mathcal{N}(0,\boldsymbol{I}). 
\end{equation}

These steps are summarized in Algorithm \ref{alg:dice}. The proposed formulation estimates the posterior distribution $p(\mathbf{x}_0|\mathbf{y})$ sampling as the agreement within data-consistency and the denoising DM. This approach differs from common posterior sampling methods in DM. For instance, in DPS \cite{DPS}, the posterior sampling is performed by projecting the generated image into the data-consistency manifold via gradient descent update. On the other hand, DiffPIR \cite{DiffPIR} includes the denosing DM into an HQS-based PnP framework, thus decoupling data-fidelity and denosing prior. Our approach provides a more flexible formulation as multiple agents promoting different properties of the target image can be incorporated into the conditioning sampling problem.

\begin{figure*}[]
    \centering
    \includegraphics[width=\textwidth]{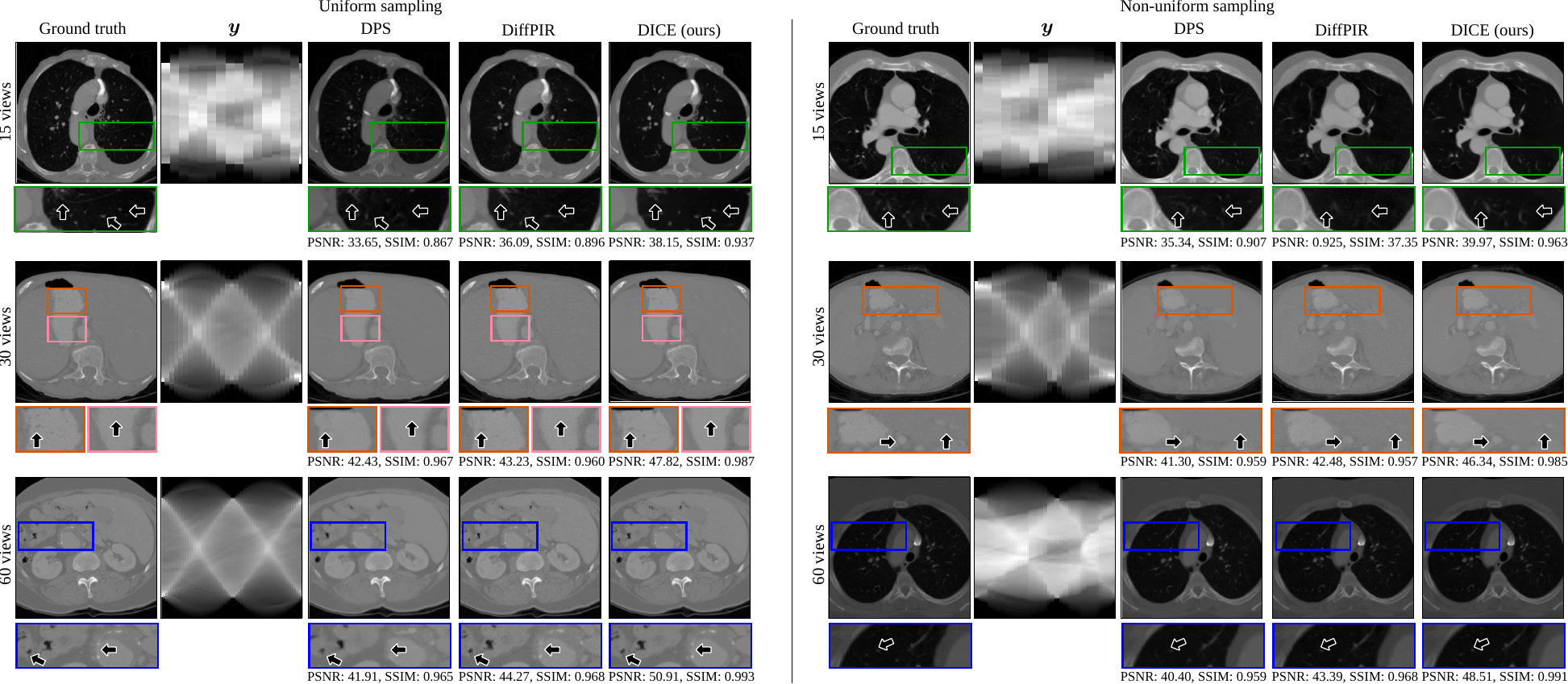}
    \vspace{-0.4cm}
    \caption{Visual comparison of CT reconstructions under uniform and non-uniform sparse-view sampling (15, 30, and 60 views), obtained with DPS, DiffPIR, and DICE.}
    \label{fig:visual_results}
    \vspace{-0.4cm}
\end{figure*}

\section{Experiments}   

\subsection{Diffusion model training} We train a DM with a cosine variance schedule ranging from $\beta_1=1\times 10 ^{-4}$ to $\beta_T=0.02$ with $T=1000$ noise steps. The DM noise estimator network $\epsilon_{\boldsymbol{\theta}}$ was optimized for $1000$ epochs with a batch size of $4$ with the AdamW optimizer \cite{ADAMW} and a learning rate of $3 \times 10 ^{-4}$. We used $10,368$ slices from the Low-Dose Parallel Beam (LoDoPaB)-CT dataset \cite{LODOPAB}, resized to $256 \times 256$ for training the DM. The test set used consist of $170$ images extracted from the test set of the LoDoPaB dataset consists of $3,553$ slices.

\textbf{Implementation Details:}  The sub-problem in $F_1$ is solved using conjugate gradient descent for $P$ iterations:
 
 \begin{equation}
     (\boldsymbol{A}^\top \boldsymbol{A}+\zeta_t\boldsymbol{I})\boldsymbol{s}=\boldsymbol{A}^\top\boldsymbol{y}+\zeta_t\boldsymbol{v}_1.
 \end{equation}
 
 All experiments were conducted on an NVIDIA RTX 4090 GPU in a PyTorch environment. The full codebase will be released at \textbf{\url{https://github.com/leonsuarez24/DICE}}.

\subsection{Reconstruction of subsampled sinograms}
%\vspace{-0.1cm}
In this experiment, we analyze quantitatively and qualitatively the performance of reconstruction subsampled sinograms. Sinograms were simulated with the parallel beam geometry with 180 views (a total of 180° with a resolution of 1°). Three uniform and non-uniform acquisition scenarios were evaluated: 15, 30, and 60 views from the total 180 views, corresponding to undersampling ratios of ${1}/{12}$, ${1}/{6}$, and ${1}/{3}$, respectively. These experiments use the parameters $\tau_1=0.5$, $\rho=0.9$, $K=5$ Mann iterations, and $P=5$ conjugate gradient descent steps. 

We compare DICE with Filtered back-projection (FBP) \cite{schofield2020image}, PnP-FISTA using the DnCNN \cite{DnCNN} as deep-denoiser (trained with the same training set of the DM), and the state-of-the-art conditioned diffusion sampling methods DPS \cite{DPS} and DiffPIR \cite{DiffPIR}. In this experiment, DPS, DiffPIR, and DICE used $T=1000$ for sampling. The proximal step of DiffPIR was solved using conjugate gradient descent with 100 iterations. To evaluate the reconstruction results, we use the quantitative metrics: peak signal-to-noise ratio (PSNR) and structural similarity index measure (SSIM).

Figure \ref{fig:visual_results} shows the reconstruction results of the DPS, DiffPIR, and DICE methods for the uniform and non-uniform acquisition settings of $15$, $30$, and $60$ views, where DICE shows significantly better and more accurate reconstructions compared to the baselines. Additionally, zoom-in sections reveal that DICE exhibits better high-frequency reconstruction details, preserving finer details. Table \ref{tab:reconstruction_results_sampling}, summarizes the average PSNR and SSIM reconstruction results in the test set of all methods, for the uniform and non-uniform acquisition scenarios with 15, 30, and 60 views. On average, DICE achieves significantly better reconstruction performance than the baselines.

\begin{table}[t!]
  \centering
  \caption{Average reconstruction PSNR and SSIM of uniform and non-uniform sparse–view CT under three sampling scenarios. 
           Best results are \textbf{bold}; second–best are \underline{underlined}.}
  \label{tab:reconstruction_results_sampling}

  \setlength{\tabcolsep}{6pt}        % horizontal padding
  \renewcommand{\arraystretch}{1.15} % vertical padding

  {%
  \begin{tabular}{l l *{6}{c}}
    \toprule
    \textbf{Sampling} & \textbf{Method} &
    \multicolumn{2}{c}{15 views $(1/12)$} &
    \multicolumn{2}{c}{30 views $(1/6)$} &
    \multicolumn{2}{c}{60 views $(1/3)$} \\
    \cmidrule(lr){3-4}\cmidrule(lr){5-6}\cmidrule(lr){7-8}
     & & \textbf{PSNR} $\!\uparrow$ & \textbf{SSIM} $\!\uparrow$ &
         \textbf{PSNR} $\!\uparrow$ & \textbf{SSIM} $\!\uparrow$ &
         \textbf{PSNR} $\!\uparrow$ & \textbf{SSIM} $\!\uparrow$ \\
    \midrule

    % ---------- Uniform ----------
    \multirow{5}{*}{\textbf{Uniform}}
    & FBP  \cite{schofield2020image}                  & 21.69 & 0.423 & 26.09 & 0.601 & 30.93 & 0.791 \\
    & PnP‐FISTA \cite{DnCNN} & 30.25 & 0.770 & 33.81 & 0.840 & 38.47 & 0.918 \\
    & DPS \cite{DPS}         & 35.95 & 0.903 & 38.93 & 0.938 & 40.24 & 0.950 \\
    & DiffPIR \cite{DiffPIR} & \underline{37.79} & \underline{0.921} & \underline{40.97} & \underline{0.949} & \underline{43.20} & \underline{0.964} \\
    \rowcolor{gray!15}
    & DICE (Ours)            & \textbf{40.51} & \textbf{0.957} & \textbf{44.92} & \textbf{0.979} & \textbf{48.51} & \textbf{0.989} \\
    \midrule

    % ---------- Non-uniform ----------
    \multirow{5}{*}{\textbf{Non-uniform}}
    & FBP  \cite{schofield2020image}                  & 19.82 & 0.386 & 23.36 & 0.549 & 26.81 & 0.691 \\
    & PnP‐FISTA \cite{DnCNN} & 29.51 & 0.756 & 33.37 & 0.841 & 37.20 & 0.908 \\
    & DPS \cite{DPS}         & 34.74 & 0.891 & 38.38 & 0.934 & 39.61 & 0.945 \\
    & DiffPIR \cite{DiffPIR} & \underline{37.25} & \underline{0.918} & \underline{40.73} & \underline{0.948} & \underline{42.95} & \underline{0.963} \\
    \rowcolor{gray!15}
    & DICE (Ours)            & \textbf{39.54} & \textbf{0.954} & \textbf{44.35} & \textbf{0.978} & \textbf{47.64} & \textbf{0.988} \\
    \bottomrule
  \end{tabular}}%
  \vspace{-0.5cm}
\end{table}

\subsection{Ablation studies}

\begin{figure}[H]
    \vspace{-0.65cm}
    \centering
    \includegraphics[width=0.75\columnwidth]{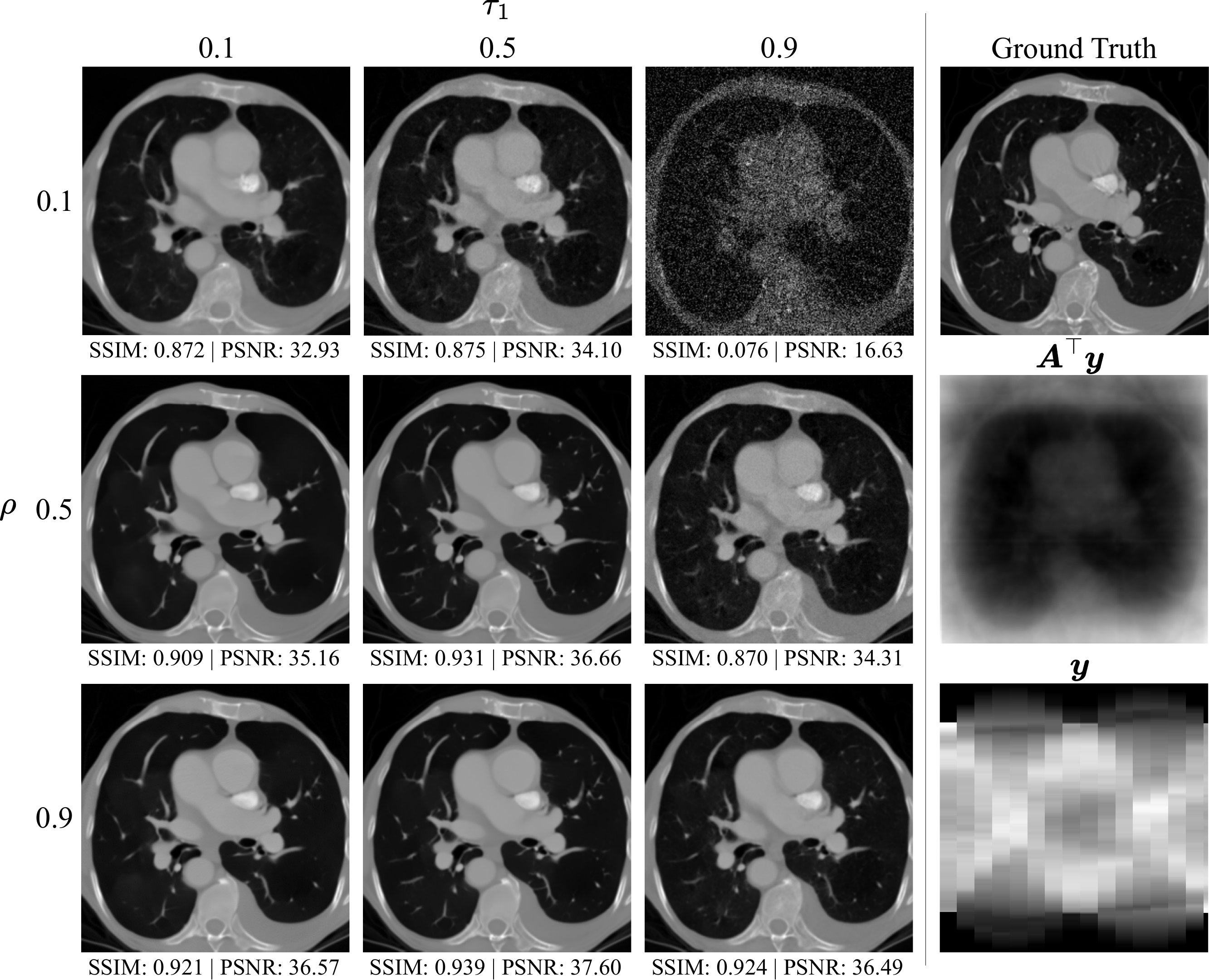}
    \vspace{-0.2cm}
    \caption{Effect of the parameters: Strength of agent $F_1$, $\tau_1$, and strength of the relaxation parameter $\rho$. }
    \label{fig:mu_tau}
\end{figure}

\textbf{Effect of $\boldsymbol{\rho}$ and $\boldsymbol{\tau_i}$:} The parameters of DICE $\rho$ controls the relaxation or strength of the CE updates, and the parameters $\tau_1$ and $\tau_2=1-\tau_1$ control the strength of the agents $F_1$ and $F_2$, respectively. To illustrate their effect, Figure \ref{fig:mu_tau} shows reconstruction results of a uniform acquisition scenario of $15$ views, with $K=5$ Mann iterations and $P=50$ conjugate gradient descent steps. From these reconstructions, it can be observed that the best reconstruction result is obtained with $\rho = 0.9$ and $\tau_1 = \tau_2 = 0.5$, which means the Mann iterations relies heavily on the current update direction, conserving few information from the previous iteration, and both agents $F_1$ and $F_2$ contribute equally to the reconstruction process, indicating balanced influence in the collaborative scheme.

\begin{figure}[H]
    \centering
    \vspace{-0.35cm}
    \includegraphics[width=0.56\columnwidth]{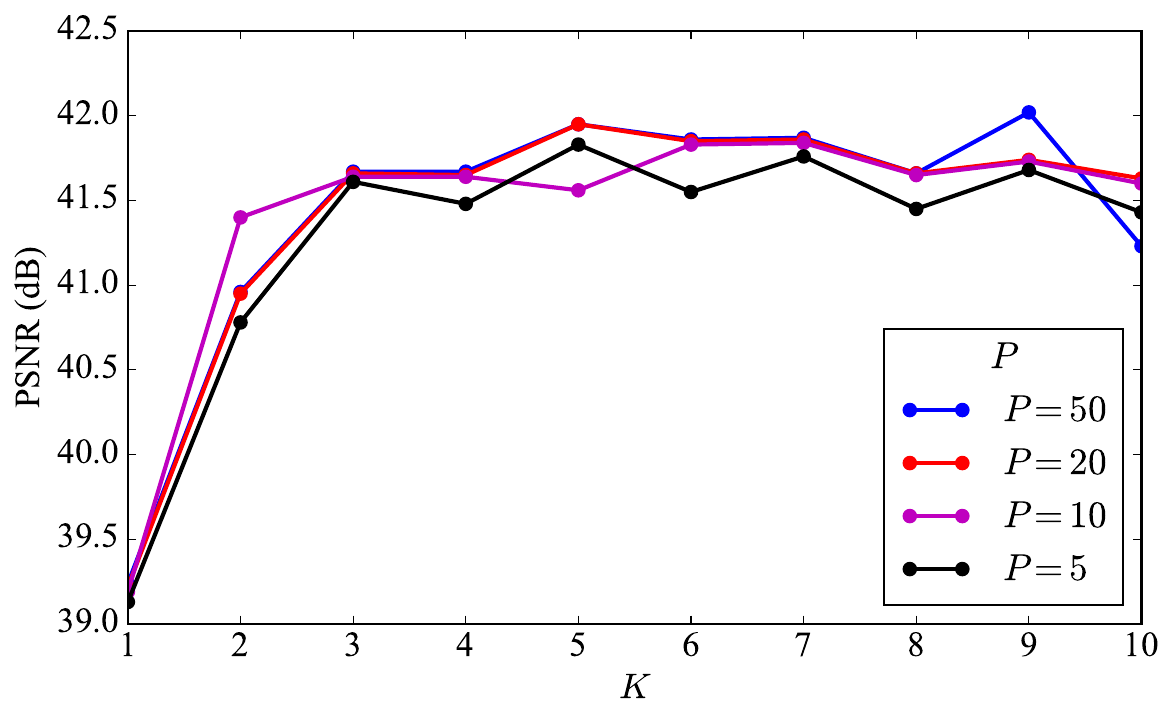}
    \vspace{-0.5cm}
    \caption{Effect of parameters: Mann iterations $K$, and conjugate gradient descent steps $P$.}
    \label{fig:K_P}
\end{figure}

\textbf{Effect of $\boldsymbol{K}$ and $\boldsymbol{P}$:} We analyze how the number of Mann iterations $K$ and conjugate-gradient steps $P$ affect reconstruction quality in Algoritm \ref{alg:dice}. Figure \ref{fig:K_P} plots the average PSNR of 20 images for different configurations of these parameters under a uniform-angle acquisition scenario with 30 views, using the parameters $\rho = 0.9$ and $\tau_1 = 0.5$. The results show that at $K = 5$ Mann iterations, the reconstruction results converge, while increasing the number of conjugate-gradient steps yields only marginal gains. Beyond $K = 5$, the PSNR oscillates, likely because the relatively large relaxation parameter $\rho$ introduces instability.

\begin{figure}[H]
    \vspace{-0.25cm}
    \centering
    \includegraphics[width=0.56\columnwidth]{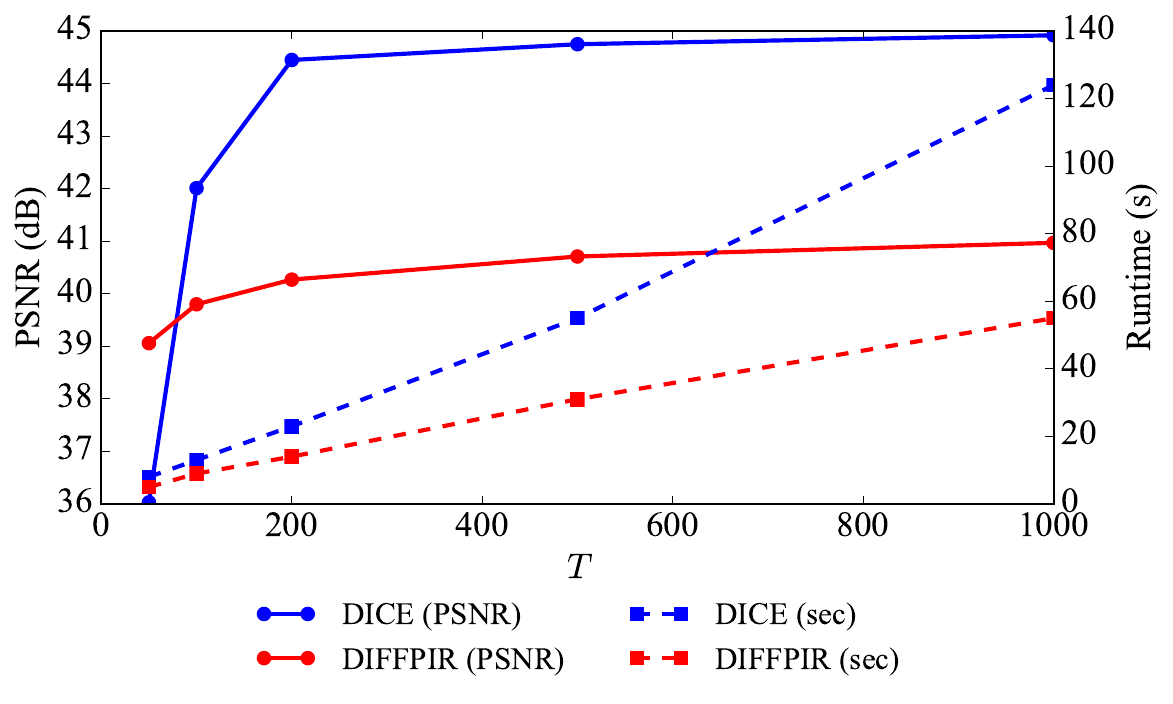}
    \vspace{-0.5cm}
    \caption{Effect of skipping time-steps}
    \label{fig:runtime}
    \vspace{-0.4cm}
\end{figure}

\textbf{Skipping time-steps:} While DICE significantly improves reconstruction performance over existing methods, it introduces additional sampling time due to solving a CE problem at each diffusion sampling step. Therefore, we analyze the performance of DICE using a reduced number of sampling steps, $T\in\{50, 100, 200, 500, 1000\}$. Figure \ref{fig:runtime} shows the average PSNR and sampling time per image for the test set images with a uniform acquisition setting of 30 views, alongside the DiffPIR baseline for each value of $T$. The results indicate that, for $T\ge100$ DICE significantly outperforms the DiffPIR baseline in reconstruction performance, with slightly increasing sampling time at timesteps  $T=100$ and $T=200$; however, at $T=50$, its performance deteriorates.

\section{Conclusions \& Future work}

This work proposed \textit{DICE: Diffusion consensus equilibrium}, a framework that casts the conditioning generative process of a DM as a consensus equilibrium problem between two agents: (i) the data-consistency agent, given by a proximal map, and (ii) the generative prior agent, given by a denoising step. Simulation with sparse-view CT with high undersampling ratios demonstrates the robustness of the proposed approach, providing high-quality reconstruction results and outperforming state-of-the-art baselines. Future work will exploit DICE’s flexibility to seamlessly integrate diverse priors (such as learned image denoisers, restoration networks, and classical sparsity models), extend the framework to additional scientific imaging inverse problems like MRI and geophysical imaging, and generalize it to other diffusion conditioning strategies.%Future work will focus on extending DICE to other scientific inverse problems, such as magnetic resonance imaging and geophysical imaging, and exploring its generalization to other diffusion conditioning methods.

%\section{Copyright}
%© 2025 IEEE. Personal use of this material is permitted. Permission from IEEE must be obtained for all other uses, in any current or future media, including reprinting/republishing this material for advertising or promotional purposes, creating new collective works, for resale or redistribution to servers or lists, or reuse of any copyrighted component of this work in other works.

%\newpage
%\balance
\bibliographystyle{IEEEtran}
\bibliography{references}

\end{document}